\PassOptionsToPackage{dvipsnames,table,xcdraw}{xcolor}
\documentclass[10pt,twocolumn,letterpaper]{article}

\usepackage[pagenumbers]{cvpr} 

\usepackage[dvipsnames]{xcolor}

\usepackage{times}
\usepackage{epsfig}
\usepackage{graphicx}
\usepackage{amsmath}
\usepackage{amssymb}
\usepackage{booktabs}
\usepackage{comment}
\usepackage{multirow}
\usepackage{subcaption}
\usepackage[table,xcdraw]{xcolor}
\usepackage{tikz}
\usepackage{pgfplots}
\usepackage{appendix}
\usepackage{balance}

\usepackage{layouts}

\newcommand{\fvec}[1]{\mathbf{#1}}

\usepackage[pagebackref,breaklinks,colorlinks]{hyperref}
\def\paperID{333} 
\def\confName{3DV\xspace}
\def\confYear{2024\xspace}

\pgfplotsset{compat=1.9}
\pgfplotsset{
    myplotstyle/.style={
    legend style={draw=none, font=\small},
    legend cell align=left,
    legend pos=north east,
    ylabel style={align=center, font=\bfseries\boldmath},
    xlabel style={align=center, font=\bfseries},
    x tick label style={font=\bfseries\boldmath},
    y tick label style={font=\bfseries\boldmath},
    scaled ticks=false,
    every axis plot/.append style={thick},
    },
}

\title{Ray-Patch: An Efficient Querying for Light Field Transformers}

\author{Tom\'as Berriel Martins 
\qquad
Javier Civera\\
I3A, University of Zaragoza, Spain\\
{\tt\small \{tberriel,jcivera\}@unizar.es}
}

\begin{document}
\maketitle
\begin{abstract}
In this paper we propose the Ray-Patch querying, a novel model to efficiently query transformers to decode implicit representations into target views. Our Ray-Patch decoding reduces the computational footprint and increases inference speed up to one order of magnitude compared to previous models, without losing global attention, and hence maintaining specific task metrics. 
The key idea of our novel querying is to split the target image into a set of patches, then querying the transformer for each patch to extract a set of feature vectors, which are finally decoded into the target image using convolutional layers. 
Our experimental results, implementing Ray-Patch in 3 different architectures and evaluating it in 2 different tasks and datasets, demonstrate and quantify the effectiveness of our method, specifically a notable boost in rendering speed for the same task metrics.
\end{abstract}    
\section{Introduction}
\label{sec:intro}

Autonomous agents rely typically on explicit representations of the environment for localization and navigation, such as point clouds \cite{campos2021orb,teed2021droid}, or voxels \cite{oleynikova2017voxblox,breyer2021volumetric}. 
However, such approaches lack topological or semantic information, struggle to generalize to changes to novel viewpoints, and do not scale properly to tasks that require reasoning about 3D geometry and affordances.
Autonomous agents require semantic, meaningful, and informative representations to properly understand and interact with their environment and perform complex tasks \cite{cadena2016past, rosen2021advances}.

Implicit representations are better suited to reasoning and are then relevant, as they capture in a continuous space the main high-level features of the scene. 
Many approaches focus on 3D geometry without topological restrictions using learned occupancy or signed distance functions \cite{chen2019learning, park2019deepsdf, michalkiewicz2019implicit, mescheder2019occupancy, peng2020convolutional, eslami2018neural}. Nevertheless, the recent success of neural fields to encode the tridimensional geometry and lighting of a scene has revolutionized the field \cite{tewari2022advances}. 
Although Neural Radiance Fields (NeRFs) focused initially on learning colour and occupancy models in a 3D space~\cite{mildenhall2020nerf}, they have demonstrated promise in a wide array of tasks such as scene segmentation \cite{zhi2021place, kundu2022panoptic, cao2022monoscene}, depth estimation \cite{guizilini2022depth}, SLAM \cite{sucar2021imap, zhu2022nice, azinovic2022neural}, scene editing \cite{gu2021stylenerf, jang2021codenerf,lazova2022control, bautista2022gaudi,stelzner2021decomposing, niemeyer2021giraffe}, and many more \cite{tewari2022advances}.
\begin{figure}
    \centering
     \newcommand\y{105}
    \begin{picture}(\linewidth,\y)(0,0)
    \put(0,14){\includegraphics[width=\linewidth]{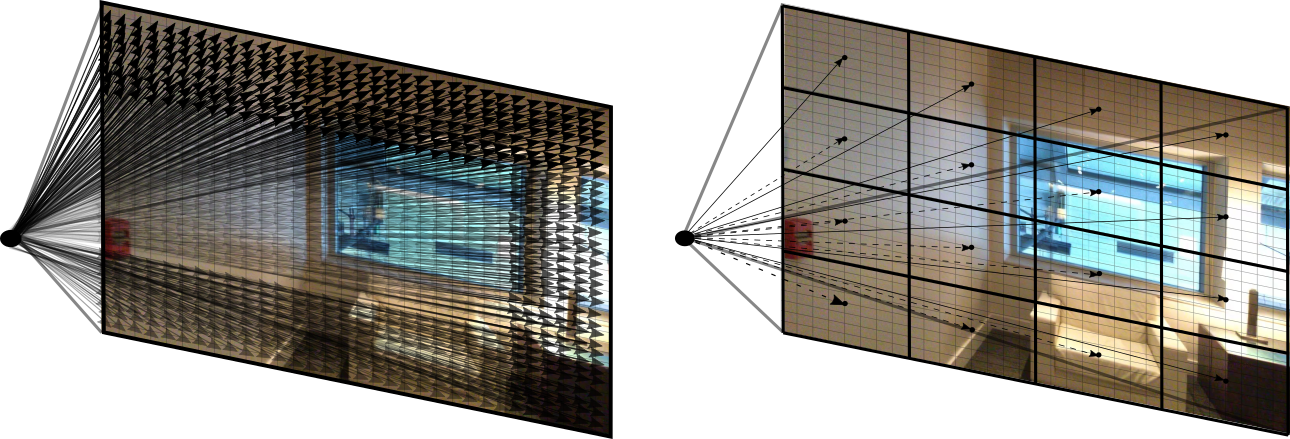}}
    \put(35, 0){\fontsize{10}{0}\selectfont Pixel querying}
    \put(130, 0){\fontsize{10}{0}\selectfont Ray-Patch querying (ours)}
    
    \end{picture}
    \caption{Light Field Networks sample a ray per pixel to render the target image (left). Our Ray-Patch (right) groups pixels in \(k\times k\) patches and samples a ray per patch, reducing the querying cost by a factor of \(k^2\) without loosing accuracy.}
    \label{fig:querying_comparison}
\end{figure}

The main limitations of neural rendering are \( 1) \) the exhaustive querying of the model that is required to recover each pixel of a specific viewpoint, and \(2)\) the need to fit the NeRF model for each scene. 
Several approaches reduce the 3D querying cost using depth \cite{ wei2021nerfingmvs, lin2022efficient, roessle2022dense, deng2022depth}, geometry \cite{trevithick2020grf, chen2021mvsnerf, yu2021pixelnerf, wang2021ibrnet}, or changing the discretization \cite{liu2020neural, yang2021learning, mueller2022instant}, and avoid per-scene optimization using latent vectors \cite{trevithick2020grf, liu2020neural, chen2021mvsnerf, yu2021pixelnerf, wang2021ibrnet, guizilini2022depth, lazova2022control}. Among them, the extensions of Light Field Networks (LFNs) \cite{sitzmann2021light} with transformers (Light Field Transformers or LFTs) \cite{sajjadi2022scene, sajjadi2022object, guizilini2022depth} have shown potential to solve both limitations, although they are constrained by the quadratic scaling of attention. Despite recent attempts to reduce it, these either modify the attention algorithm for a less expensive but less effective version \cite{xiong2021nystromformer, tay2020long, zaheer2020big}; or are based on extensive optimization of last generation hardware and software \cite{dao2022flashattention, dao2023flashattention2}. 
Therefore, despite significant advances in both qualitative performance and efficiency, all these approaches are still far from being scalable to real scenarios with real-time performance. 

In this work we propose Ray-Patch, a novel decoding method that reduces the computation and memory load of LFTs up to one and two orders of magnitude respectively, while keeping the quality of the output. We developed Ray-Patch as a generic decoder that can be implemented on any LFT architecture in the literature. Instead of the typical per-pixel querying, we group all pixels in a square patch, as shown in \cref{fig:querying_comparison}, and compute a set of feature vectors, which are then grouped and decoded into the target viewpoint. Specifically, it combines a transformer decoder with convolutional neural networks to reduce the cost of the decoder processing. This results in a drastic reduction in the number of queries, which impacts quadratically in the cost, allowing to decode high-resolution images while keeping and sometimes even improving the training convergence.

\section{Related Work}
\label{sec:related}
A NeRF \cite{mildenhall2020nerf} is an implicit representation of a scene that is learnt from a sparse set of multiple views of such scene, annotated with their corresponding camera poses. NeRFs encode a continuous volumetric model of a scene that can be used to render photorealistic novel views from arbitrary viewpoints.
The rendering process involves projecting pixels into rays, sampling 3D positions along the rays, and querying a Multilayer Perceptron (MLP) network that predicts the colour and occupancy of the sampled 3D points.
Despite its versatility and impressive results in various applications~\cite{tewari2022advances}, NeRFs suffer from two major limitations: exhaustive 3D sampling is required to decode each pixel, and a new model must be trained for each new scene.
\paragraph{Multi-scene implicit representations.}
One of the most promising approaches to enable the generalization of neural fields across multiple scenes is conditioning the output of the MLP to a latent vector that is optimized for each scene at test time. NSVF \cite{liu2020neural} discretizes the 3D space into a sparse voxel octree associating each voxel with a feature vector that guides the sampling of 3D points. Control-NeRF \cite{lazova2022control} also utilizes voxel features, but employs a multi-resolution incremental training of the full feature volume. Nice-SLAM \cite{zhu2022nice} leverages a multi-resolution feature grid to encode the scene while simultaneously performing camera tracking. InstantNGP \cite{mueller2022instant} implements multi-resolution voxelization as a hash encoding, where the MLP is responsible for avoiding hash collisions, resulting in remarkable improvements in reconstruction quality and convergence.

Other approaches involve using an encoder architecture to compute latent vectors, and use these to condition the NeRF decoder. GRF \cite{trevithick2020grf} projects sampled 3D points into feature maps of the input views computed with a CNN encoder-decoder. These are first processed by shared MLP to condition on the sampled 3D point, and then aggregated using an attention module. The final feature vector is fed to a final MLP to estimate the 3D point colour and density.
PixelNeRF \cite{yu2021pixelnerf} extends this approach by adding the feature vector as a residual in each layer of the MLP and using a simple average pooling instead of an attention module. IBRNet \cite{wang2021ibrnet} also projects 3D points into nearby views to estimate the conditioning feature vectors, although it relies on a differentiable rendering algorithm rather than a neural representation to estimate the final colour and depth. 
MVSNeRF \cite{chen2021mvsnerf} proposes the use of a CNN and homography reprojections to build a cost volume and compute the features. ENeRF \cite{lin2022efficient} builds on MVS approaches to build the cost volume, guide the sampling, and condition the reconstruction. However, these methods that rely on homographies are limited to a small range of views in front of the reference camera, require accurate camera poses, and are not robust to occlusions.

SRT \cite{sajjadi2022scene} introduced a Transformer \cite{vaswani2017attention} to encode and decode the scene, performing self-attention between the features of different points of view. It generates a latent representation of the scene, which is decoded using a light-field cross-attention module. OSRT \cite{sajjadi2022object} improves SRT by disentangling the object of the scene and improving its control. DeFiNe \cite{guizilini2022depth} replaces the basic Transformer for a PerceiverIO \cite{jaegle2021perceiver} reducing the cost of self-attention and scaling to bigger resolutions. Despite the low rendering time achieved for new scenes and novel points of view, these methods computation cost scales poorly due to the use of attention. Consequently, they may not be suitable for large scenes with high-resolution images.
\begin{figure*}[t]
\begin{subfigure}{0.48\linewidth}
\includegraphics[width=\linewidth]{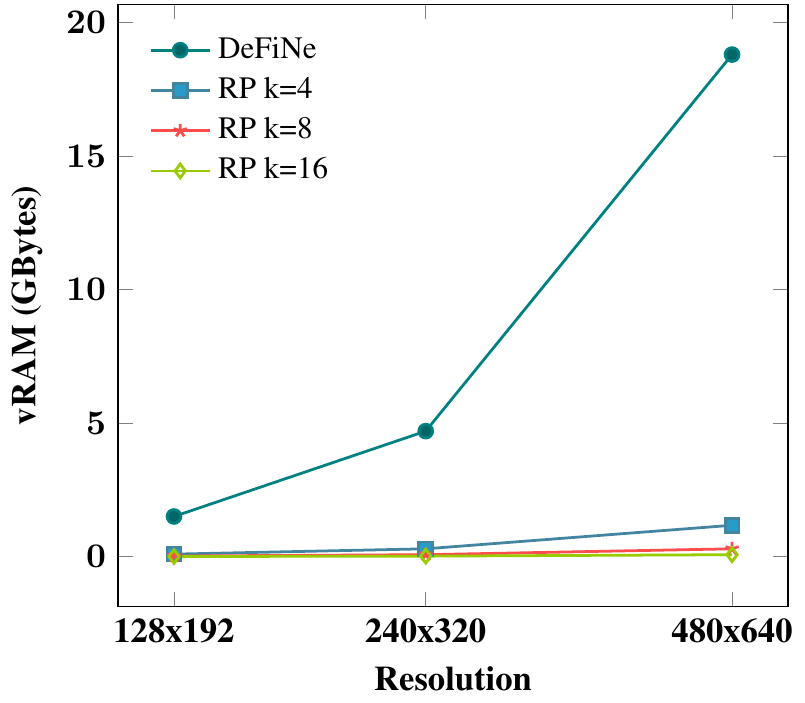}
\caption{}
\label{fig:ramVsRes_define}
\end{subfigure}
\begin{subfigure}{0.49\linewidth}
\vspace{-0.5cm}
\includegraphics[width=\linewidth]{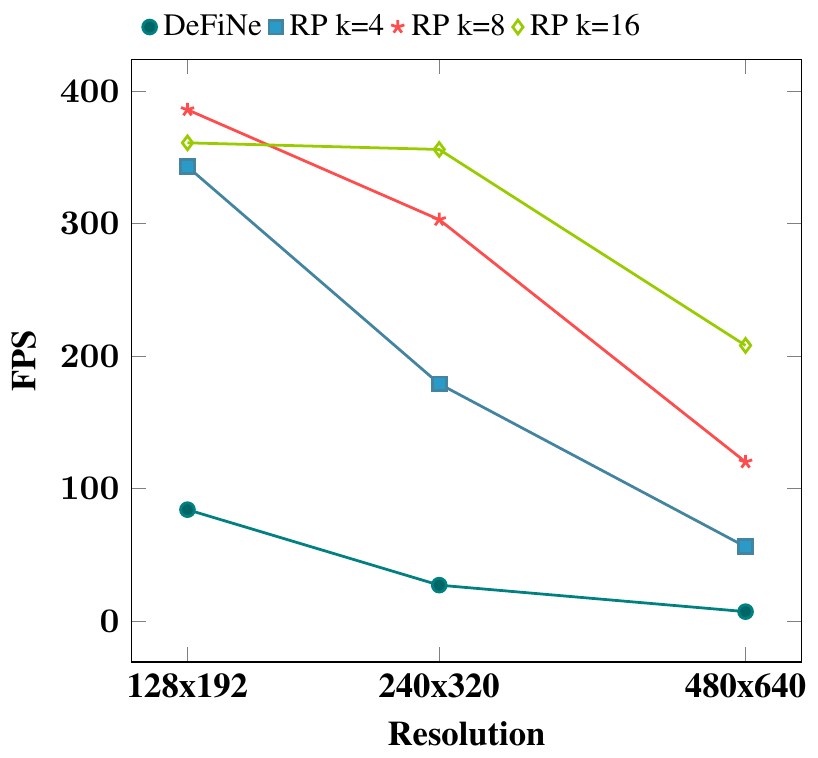}
\caption{}
\label{fig:fpsres}
\end{subfigure}
\caption{\textit{(a)} \textbf{Peak GPU vRAM usage due to attention} for decoding a single image. vRAM usage scales linearly with the number of pixels (quadratically with resolution). Ray-Patch querying reduces \(\times10\) required resources on standard resolutions. Note that $x$-axis is in logarithmic scale. \textit{(b)} \textbf{Single image rendering speed scaling.} The use of the Ray-Patch decoder increase rendering speed at high resolutions up to real-time for DeFiNe. To keep a fixed rendering speed, the patch size should increase at the same pace as the number of pixels.}
\end{figure*}
\paragraph{Rendering from implicit neural representations.}
To render a pixel, NeRF evaluates 3D coordinates sampled along a ray combining both uniform and stratified distribution. This random sampling results in a great number of evaluation wasted on empty space. 
DONeRF \cite{neff2021donerf} trains an oracle network supervised on dense depth map on synthetic data. Although it achieves outstanding results, the setup is hard to generalize to real environments.
NerfingMVS \cite{wei2021nerfingmvs} trains a monocular depth estimation network, supervised with sparse depth from Structure from Motion (SfM), to guide the sampling and reduce empty queries both at training and test. Roessle \etal \cite{roessle2022dense} instead uses a depth completion network to estimate depth and uncertainty from the sparse SfM prior at training time, improving performance while still requiring multiple sampling at test time. In contrast, ENeRF \cite{lin2022efficient} uses an estimated cost volume to predict the depth and guide the sampling without any explicit depth supervision nor structure from motion.
Other alternative like Neural RGB-D \cite{azinovic2022neural} and Mip-NeRF RGB-D \cite{dey2022mip}, directly use RGB-D sensor as prior for the depth sampling.
Despite reducing the number of samples, most of these approaches still perform multiple samples per pixel due to the error range of depth data and estimation.
Instead, Light Field Network \cite{sitzmann2021light} directly evaluate the unprojected pixels, parameterized as a ray, reducing the model query to the number of pixels. Although, this approach is able to perform real-time rendering of novel views without requiring heavy optimizations, it does not generalize yet to high resolution scenes.

\section{Preliminaries: Light Field Transformers}
\label{sec:preliminaries}
While NeRFs learn a scene representation associated to a continuous space of 3D points, Light Field Networks (LFNs) \cite{sitzmann2021light} rely on 3D rays parametrized with Plücker coordinates to learn similar representations. This subtle difference reduces significantly the cost to decode a view of the scene, from several samples to a single sample per pixel. Despite this, LFNs are limited to simple settings, do not enforce geometric consistency and their ray parameterization is not robust to occlusions.

Light Field Transformers (LFTs) are an extension of LFNs which use a transformer architecture, a ray parametrization robust to occlusions, enforce geometric consistency through the training procedure, and encode and decode points of view of a scene without per-scene optimization \cite{sajjadi2022scene,sajjadi2022object,guizilini2022depth}.  

\subsection{Transformers}
Transformers \cite{vaswani2017attention} are deep encoder-decoder neural models that incorporate attention mechanisms in their architecture. 
The encoder first performs self-attention on a set of tokens to extract common features. 
The decoder then uses cross-attention between the extracted features and a set of queries to generate an output per query. 
The attention block consists of a Multi-Head Attention (MHA) layer, followed by a Feed-Forward (FF) layer, with a skip-connection and layer normalization after each of them.
In each head \(h\), MHA operates in parallel a Scaled Dot-Product attention 
\begin{equation}
    \text{Attention}_h(Q,K,V) = \text{softmax}\left(\frac{QK^T}{\sqrt{d_k}} \right)V,
    \label{eq:attention}
\end{equation}
over a set of the three inputs: keys~(\(K\)), values~(\(V\)), and queries~(\(Q\)).
Each head linearly projects the inputs to reduced dimensions, \(d_k\) for \(Q\) and \(K\) and \(d_v\) for \(V\), performs the attention operation, and then projects the output back to its original dimension.
To perform self-attention \(Q=K=V\) are the tokens to encode. Instead for cross-attention \(K=V\) are the extracted features, while \(Q\) is the queries to decode.  

\paragraph{Computational complexity.} Linear projections have a complexity of \(\mathcal{O}(n d_0 d_p )\) with \(n\) the length of the sequence and \(d_0 \text{ and } d_p\) the dimensions before and after the projection.
Instead, the scaled-dot product has \(\mathcal{O}(n_q n_{kv} d_k)\) complexity, being \(n_q \text{ and } n_{kv}\) the number of queries and keys/values respectively. For self-attention, \(n_q = n_{kv}\) and then the complexity is \(\mathcal{O}(n_{q}^2 d_k)\).

\subsection{Scene Representation Transformer}
\begin{figure*}
    \centering
    \includegraphics[width=\linewidth]{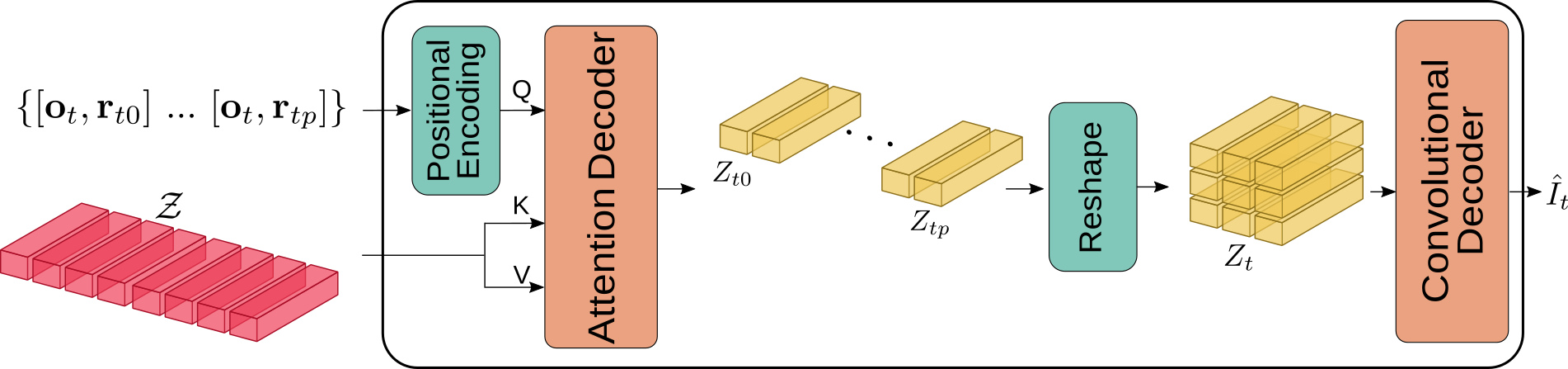}
    \caption{\textbf{Ray-Patch querying.} Given a latent representation of a scene  \(\mathcal{Z}\), in order to render an image \(I_t\) of shape \(h\times w\), a query is performed for each patch \(p\). Each patch is parametrized with a ray \(\mathbf{r}_{tp}\)  that passes through it and the the camera position \(\mathbf{o}_t\). 
The queries are encoded with multiple Fourier frequencies, and fed to the attention decoder to compute a feature vector per query.
The feature vectors of the image are re-shaped as a rectangle and forward-passed through the convolutional decoder to obtain the target image render \(\hat{I}_t\)}
    \label{fig:decoder}
\end{figure*}
 The Scene Representation Transformer (SRT) \cite{sajjadi2022scene} is an encoder-decoder LFT, which parametrizes rays with its 3D coordinates and their origin position. Given a set of \(N\) input views \(\{I_n\}\)\footnote{We abuse notation here for simplicity, \(\{\circ_n\} \equiv \{\circ_1, \hdots, \circ_N\}\)}, and their relative camera poses \(\{P_n\}\) with camera Instrinsic parameters \(\{K_n\}\), the encoder \(\mathcal{E}\) generates a set-latent scene representation (SLSR) 
\begin{equation}
    \mathcal{Z}= \mathcal{E}\left(\{I_n, P_n \}\right),
\end{equation}
To decode a view of the scene, the light-field based decoder is queried. Each query refers to the ray direction and camera center for a given pixel, and recovers its RGB values. To decode a full view, as many queries as pixels are needed.

The encoder is made of two parts. First, a convolutional network extracts features from the scene images. Then a set self-attention blocks computes common features between the multiple views of the scene to generate a SLSR. The decoder is a two-blocks cross-attention module. It performs attention between the ray queries and the SLSR to generate the RGB pixel values.
SRT has been extended by OSRT \cite{sajjadi2022object} to disentangle its latent representation by integrating it with Slot-Attention~\cite{locatello2020object} and designing the Slot Mixer Decoder. Using Slot Mixer attention weights, OSRT is able to generate unsupervised segmentation masks.
\paragraph{Attention cost.} 
With a convolutional encoder which halves the resolution (divides by four the number of queries) three times, \(n_{q} = n_{kv} = N \frac{h \times w}{64}\) for the encoder self-attention block. Therefore the complexity is 
\begin{equation}
   \mathcal{O}\left(\left(\frac{N hw}{64}\right)^2 d_k\right).
\end{equation}
Instead, for the decoder cross-attention block to decode an image, \( n_{q} = h\times w \) and \(n_{kv} = N\frac{h \times w}{64}\), therefore the complexity is 
\begin{equation}
   \mathcal{O}\left(\frac{N \left({hw}\right)^2}{64} d_k\right).
\end{equation}
Doubling the resolution will increase by a factor of 4 the total number of pixels \(h\times w\), and by 16 the computational complexity.
As a consequence, both SRT and OSRT are limited due to the quartic scaling of the attention cost with respect to the resolution of the images, and to the quadratic cost with respect to the number of input images \(N\).
\subsection{Depth Field Network}
The Depth Field Network (DeFiNe) \cite{guizilini2022depth} can be considered as an extension of SRT. As its main novelties, the convolutional encoder is a pretrained ResNet-18, the cross-attention decoder is reduced from two blocks to one, and a set of geometric data augmentations are proposed for stereo and video depth training. 
The main contribution is the use of a PerceiverIO \cite{jaegle2021perceiver} instead of the self-attention encoder to use a SLSR with a fixed size \(n_l\).
\paragraph{Attention cost.} With \(n_{kv} = n_l\) in both the encoder and decoder, the quadratic scaling with respect to the resolution of the images is reduced to
\begin{equation}
   \mathcal{O}\left(\frac{N hw}{64}n_l d_k\right)
\end{equation}
for the encoder attention process, and to
\begin{equation}
   \mathcal{O}\left(hw n_l d_k \right),
\end{equation}
for decoding an image. These improvements reduce the cost considerably for \(hw >> n_l\), although there is still a quadratic dependence with the number of pixels (quartic with resolution) that limit the model's use.
\section{Our Method: The Ray-Patch Decoding}
\label{sec:method}
We propose the Ray-Patch querying to attenuate the quartic complexity of Light Field Transformers with respect to image resolution. Instead of using a ray to query the cross-attention decoder and generate a pixel value, we use a ray to compute a feature vector of a square patch of pixels. Then a transposed convolutional decoder unifies the different patches' feature vectors and recovers the full image. Our approach reduces the number of queries to \(\frac{hw}{k^2}\) and the cross-attention cost by the same factor.
\paragraph{Parametrization.} To decode a target view \(I_t \in  \mathbb{R}^{h\times w\times c}\) of the scene, the view is split into  \(\frac{hw}{k^2}\) square patches of size \(\left[ k, k \right] \), being the split image now defined as \(\{I_{tp} \in \mathbb{R}^{\frac{h}{k}\times \frac{w}{k} \times 3}\}\). Each patch \( p\) is parametrized by the location of the camera \(\mathbf{o}_t\), and the ray \(\fvec{r_{tp}}\) that passes both by the camera position and the center of the patch. 
Given the camera intrinsic \({K_t}\) and extrinsic parameters \({{}^{W}{T}^{C_t}} = \left[ R_t | \fvec{o}_t \right] \in SE(3)\), the ray \(\fvec{r_{tp}}\) is computed as the unprojection of the center of patch \( p\) in the 2D camera plane.
For each patch center in homogeneous coordinates \(\fvec{x}_{tp} = \left(u_{tp},v_{tp}, 1 \right)^T\), it is first unprojected in the the camera reference frame \(C_t\),  
\begin{equation}
    \fvec{r}_{tp}^{C_t} = K_n^{-1} \cdot \fvec{x}_{tp} = \left[x_{tp}/z_{tp}, y_{tp}/z_{tp}, 1, 1 \right]^T,
\end{equation}
and after that it is translated to the world reference \(W\),
\begin{equation}
    \fvec{r}_{tp}^{W} = {}^{W}{T}^{C_t} \cdot  \fvec{r}_{tp}^{C_t}.
\end{equation}
Using Fourier positional encoding \cite{mildenhall2020nerf}, the parametrization of each patch is mapped to a higher frequency, to generate a set of queries for the decoder.
\begin{equation}
    \{\mathcal{Q}_{tp}\} = \{\gamma\left( \mathbf{o}_t\right) \oplus \gamma\left(\mathbf{r}_{tp}\right)\}
\end{equation}
\paragraph{Decoder.} The decoder \(\mathcal{D}\) is a composition 
\begin{equation}
    \mathcal{D} = \left( \mathcal{D}_{\text{CNN}} \circ \mathcal{D}_A \right)
\end{equation}
of an attention decoder \(\mathcal{D}_{A}\), followed by a convolutional decoder block \(\mathcal{D}_{\text{CNN}}\). 
The attention decoder performs cross-attention
between the queries \(\{\mathcal{Q}_{tp}\}\) and the SLSR \(\mathcal{Z}\), to compute a set of feature vectors 
\begin{equation}
    \{Z_{tp}\} = \mathcal{D}_A \left(  \{\mathcal{Q}_{tp}\},  \mathcal{Z} \right)
\end{equation}
with dimension \(f\). These vectors  ensemble a feature map \(Z_{t} \in \mathbb{R}^{\frac{h}{k}\times \frac{w}{k} \times f}\), which is decoded by the convolutional decoder into the target image
\begin{equation}
\hat{I}_t = D_{\text{CNN}}\left(Z_t\right),
\end{equation}
as shown in ~\cref{fig:decoder}.
We use a vanilla convolutional decoder \(\mathcal{D}_{CNN}\) based on GIRAFFE’s decoder~\cite{niemeyer2021giraffe}. It is a combination of upsampling blocks with convolutions and preliminary outputs.
The number of channels of the output, \(c\), will vary depending on the desired task, \eg \(c=3~\)for RGB colour image, or \(c=1~\)for depth estimation.
\begin{figure}
\input{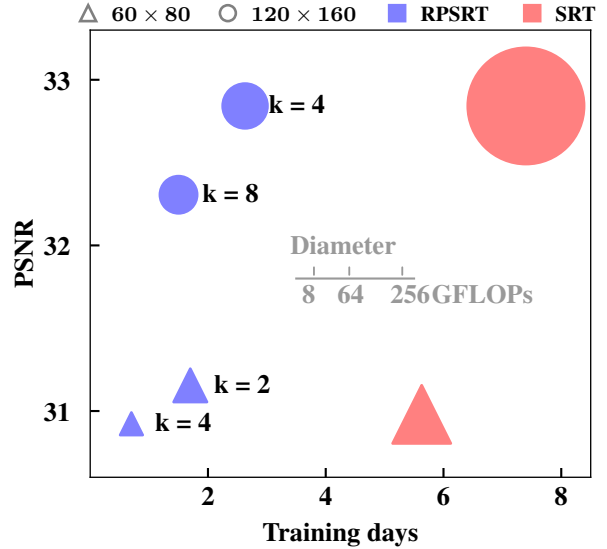}
\caption{\textbf{Training time on one V100 comparison.} For both \(60\times80~(\bigtriangleup)\) and \(120\times160~(\circ)\) resolution, Ray-Patch~(\textit{blue}) configurations, \(k=\{2,4\}\) and \(k=\{4,8\}\) respectively, achieve similar or better rendering performance than SRT~(\textit{red}), with \(60-70\%\) cost reduction.}
\label{fig:msn_psnr_vs_days}
\end{figure}
\paragraph{Integration.} The simplicity of the Ray-Patch querying allows to easily integrate it in LFTs like SRT, OSRT, or DeFiNe. Changing the number of channels of the output of their decoders to \(f\), they can be used as \(\mathcal{D}_{A}\) to decode the final image as 
\begin{equation}
\hat{I}_t = D_{cnn}\left( D_{A} \left(  \{\mathcal{Q}_{tp}\},  \mathcal{Z} \right ) \right).
\end{equation}
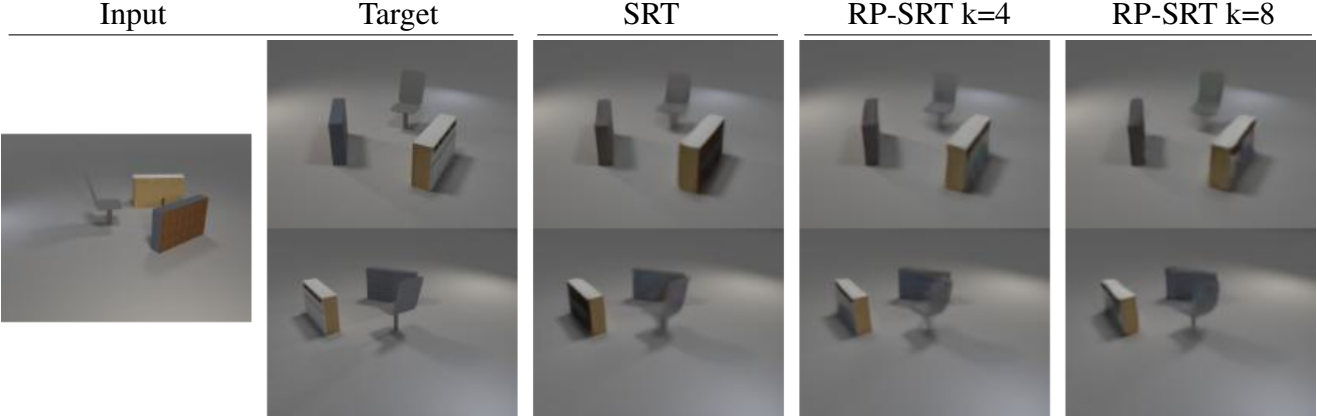
\begin{figure*}
    \centering
    \newcommand\y{158}
    \begin{picture}(\linewidth,\y)(0,0)
    \put(0,0){\includegraphics[width=\linewidth]{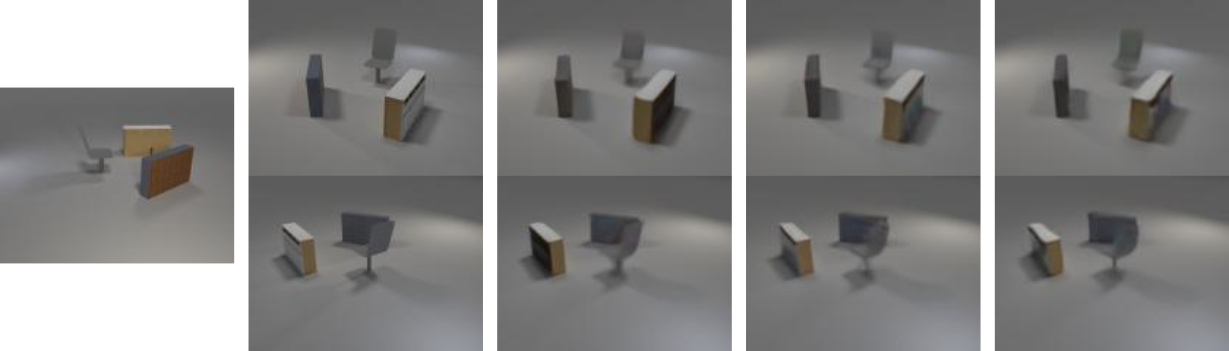}}
    \put(37, \numexpr\y-10){\fontsize{12}{0}\selectfont Input}
    \put(135, \numexpr\y-10){\fontsize{12}{0}\selectfont Target}
    \put(0, \numexpr\y-14){ \line(1, 0){191} }
    
    \put(235, \numexpr\y-10){\fontsize{12}{0}\selectfont SRT}
    \put(200, \numexpr\y-14){ \line(1, 0){91} }
    
    \put(320, \numexpr\y-10){\fontsize{12}{0}\selectfont RP-SRT k=4}
    \put(420, \numexpr\y-10){\fontsize{12}{0}\selectfont RP-SRT k=8}
    \put(301, \numexpr\y-14){ \line(1, 0){192} }
    
    \end{picture}
    
    \caption{\textbf{Novel view synthesis results on MSN-Easy.} Given an input image, the models are queried to decode target images at 120º (first row) and 240º (second row). Both SRT and Ray-Patch models (RP-SRT \(k=\{4,8\}\)) encode a coherent representation, with slight differences on colour and edges. For RP-SRT it can be seen how the bigger the patch size the more diffuse the image looks.}
    \label{fig:msn_easy}
\end{figure*}
The optimization process does not change. The model's parameters \(\theta\) are optimized on a collection of images from different scenes minimizing the Mean Squared Error (MSE) of the generated novel-views for RGB images
\begin{equation}
    \mathcal{L}_{rgb} = \frac{1}{hw} \sum_{ij} \left(\hat{I}_{t} - I_{t}\right)^2,
\end{equation}
and minimizing the absolute log difference for depth maps 
\begin{equation}
    \mathcal{L}_d = \frac{1}{n_{tp}}\sum_{tp} | \log{\hat{\mathbf{D}}_{tp}} - \log{\mathbf{D}_{tp}}|,
\end{equation}
Depth optimization is performed only over the subset \(\{tp\} \subset \{ij\} \) of target pixels with depth info, giving freedom to the model to generalize to unseen parts.
\paragraph{Attention cost.}
The proposed Ray-Patch querying reduces the complexity of the decoders to 
\begin{equation}
   \mathcal{O}\left(\frac{N \left({hw}\right)^2}{64 k^2} d_k\right),
\end{equation}
for models with the basic Transformer, like SRT and OSRT; and to 
\begin{equation}
   \mathcal{O}\left(\frac{hw}{k^2} n_l d_k\right),
\end{equation}
for PerceiverIO based models, like DeFiNe. 
Although there is still a quadratic dependency on the resolution, the attenuation introduced by the Ray-Patch querying can reduce the number of queries in up to two orders of magnitudes for high resolutions.
\section{Experimental Results}
\label{sec:results}
We evaluate Ray-Patch using two setups of different complexity.
Firstly, we integrate Ray-Patch into both SRT and OSRT for novel view synthesis on the MulstiShapeNet-Easy (MSN-Easy) dataset.
Given input images, the model encodes a representation of the scene, and its goal is decoding the other two viewpoints. In this dataset we asses the impact of different patch sizes at different resolutions on SRT implementation, and its integration on OSRT.
After that, we evaluate its ability to generalize to more challenging scenes and textures in a stereo depth task.

Secondly, we also implemented Ray-Patch into DeFiNe and evaluated on ScanNet. Given two images, the model encodes a representation, and the goal is recovering RGB and depth from the same point of view.
Following Sajjadi \etal~\cite{sajjadi2022scene, sajjadi2022object}, rendered views are benchmarked with PSNR, SSIM, and LPIPS; and segmentation masks with FG-ARI. Following Guizilini \etal~\cite{guizilini2022depth}, depths are benchmarked with Absolute Relative Error (Abs.Rel), Square Relative Error (Sq.Rel) and Root Mean Square Error (RMSE). 
Computational aspects are evaluated measuring peak RAM usage, image rendering speed as in Sajjadi \etal~\cite{sajjadi2022scene}, training time, and Float Point Operations (FLOPs) needed to encode and render an image. We assume the use of float-32 data, and report time metrics from a GPU NVIDIA Tesla V100. 
Further design and implementation details are provided on the supplementary material.
\begin{table*}
\centering
\resizebox{0.9\linewidth}{!}{
\begin{tabular}{lcccrcccrccr}
\hline
      & \multicolumn{7}{c}{MSN-Easy}                      &  \\ \cline{2-12} 
      & \multicolumn{3}{c}{\(60\times80\)} & &  \multicolumn{3}{c}{\(120\times160\)}  & & \multicolumn{2}{c}{\(120\times160\)}                    &  \\ \cline{2-4} \cline{6-8}  \cline{10-11}
      & \multirow{2}{*}{SRT} & \multicolumn{2}{c}{RP-SRT} && \multirow{2}{*}{SRT} & \multicolumn{2}{c}{RP-SRT} && \multirow{2}{*}{OSRT} & \multicolumn{1}{c}{RP-OSRT} &\\ \cline{3-4} \cline{7-8} \cline{11-11}
      &                   & k = 2   & k = 4 & &   & k = 4   & k = 8  & & & k = 8 & \\ \cline{2-4} \cline{6-8}  \cline{10-11} 
\(\uparrow\) PSNR  &  30.98 & \textbf{31.16} & 30.92 &&  \textbf{32.842} & 32.818 & 32.306 & &30.95&\textbf{31.03}&\\
\(\uparrow\) SSIM &  0.903 & \textbf{0.906} & 0.901 &&  0.934 & \textbf{0.935} & 0.929 & &\textbf{0.916}&0.915&  \\ 
\(\downarrow\) LPIPS &  0.173 & \textbf{0.163} & 0.175 &&  \textbf{0.250} & 0.254 & 0.274 & &\textbf{0.287}&0.303&\\
\(\uparrow\) FG-ARI &-&-&-&&-&-&-&&\textbf{0.958} &0.914& \\
\cline{1-11}  
\(\downarrow\) Training time & 5.6 days  & 1.7 day & \textbf{0.7 days}  && 7.4 days  & 1.7 days & \textbf{1 day}  &&25 days&\textbf{3.7 days}&\\ \cline{1-11} 
\(\downarrow\) Giga FLOPs & 48.2 & 15.8 &  \textbf{7.3} && 192.1 & 28.5 &  \textbf{19.7} & &278.6&\textbf{24.7}&\\
\(\uparrow\) Rendering speed & 117 fps & 288 fps & \textbf{341 fps} && 30 fps & 275 fps & \textbf{305 fps} & &21 fps&\textbf{278 fps}&\\\hline

\end{tabular}}
\caption{\textbf{Quantitative results on MSN-Easy.} Evaluation of new scene novel view synthesis and computational performance on a simple dataset. While SRT's performance is surpassed only by the configuration with patch size \(k=2\), Ray-Patch increases \(\times3\) and \(\times10\) the rendering speed with minimum impact.}
\label{tab:msneasy}
\end{table*}
\subsection{Datasets}
\textbf{MulstiShapeNet-Easy}~\cite{stelzner2021decomposing} has 70K training scenes and 10K test scenes with resolution \(240\times320\). Due to the high cost of training both SRT and OSRT, we work at \(60\times80\) and \(120\times160\).  
In each scene there are between 2 and 4 objects of 3 different classes: chair, table, or cabinet. The object shapes are sampled from the ShapeNetV2 dataset \cite{chang2015shapenet}.
Each scene has 3 views sampled at $120^\circ$ steps on a circle around the center of the scene, with extrinsics and intrinsics camera annotations.
For each training step, one image is used as input and the other two are used as target to be reconstructed. 

\textbf{ScanNet}~\cite{dai2017scannet} is a collection of real indoor scenes with RGB-D and camera pose information. It has ~1.2K different scenes with a total of 90K views.
We follow DeFiNe's \cite{guizilini2022depth} stereo setup: RGB input images are downscaled to a resolution of \(128\times192\); and a custom stereo split is used \cite{kusupati2020normal}, resulting in 94212 training and 7517 test samples.
\subsection{Computational performance}
While our Ray-Patch querying still has quadratic scaling with \(n_q\), the reduction we achieve in the number of queries results in a notable boost in rendering speed, as can be seen in \cref{tab:msneasy} and \cref{fig:fpsres}.
Furthermore, when increasing the resolution the patch can also be increased, keeping an appropriate rendering speed at higher resolutions.
Comparing rendering speeds for different patches and resolutions in \cref{fig:fpsres}, it can be observed how the improvement tends to saturate for big patch sizes.
As a consequence of reducing the number of queries, its impact on the scaled-dot product complexity will be out-weighted by \(n_{kv}\). For \(n_q << n_{kv}\), \(n_{kv}\) will set a minimum cost and increasing the patch size over this limit will not  be reflected on the rendering speed.
It is also worth of attention that the biggest patch does not have the lower rendering time. When \(n_q << n_{kv}\), increasing the patch size also adds more convolutions and interpolations to the convolutional decoder, hence increasing the deconvolutional overhead without reducing the cost of the attention decoder. 
Finally, the decrease in \(n_q\) implies a smaller memory peak in the softmax of the decoder attention, see \cref{fig:ramVsRes_define}. This matrix is \(n_q \times n_{kv}\). As an illustrative example, for DeFiNe, decoding a single \(960\times1280\) image, with \(n_{kv}=2048\), requires 75 GBytes of GPU memory, almost two full A100 GPUs. Instead, for the Ray-Patch querying with \(k=16\), it is reduced to only 0.3 GBytes. This notable reduction allows to increase parallelization, improving even more the rendering speed for scene reconstruction tasks.
\subsection{Novel view synthesis}
On MSN-Easy, for SRT we evaluate two different patch sizes for each resolution:
\(k=\{2,4\}\) for \(60\times80\); and \(k=\{4,8\}\) for \(120\times160\). Instead for OSRT we only evaluate at \(120\times160\) with a patch size \(k=8\).

As reported in \cref{tab:msneasy} and \cref{fig:msn_easy}, the experiment metrics for RP-SRT shows that the size of the patch impacts on the model, with smaller patches having better rendering quality on both resolutions. For smaller patches, the first decoder focus attention on less pixels than for a bigger patch, each feature vector is up-sampled less, and more information is recovered from the same amount of data. Therefore, excessively increasing the patch reduces the quality of reconstructed views, as shown by RP-SRT with \(k=8\) for \(120\times160\), which slightly underperforms the baselines's PSNR (32.3 vs 32.8). 
Nevertheless, Ray-Patch querying is still able to match rendering quality of both SRT and OSRT at \(120\times160\), with \(k=4\) and \(k=8\) respectively; and outperform at \(60\times80\), with \(k=2\). Furthermore, for similar performance our approach improves rendering speed \(\times 10\) for the highest resolution (275 vs 30 fps, and 278 vs 21 fps), and reduces training time almost \(\times 4\) (see \cref{fig:msn_psnr_vs_days} and \cref{tab:msneasy}). This is thanks to scaling the attenuation factor \textit{k} together with resolution, compensating for the increasing number of queries.
Regarding RP-OSRT's unsupervised segmentation, we up-sample the \(\frac{120}{8}\times\frac{160}{8}\) attention weights of the Slot Mixer Decoder to generate a \(120\times160\) segmentation map, achieving only slightly worse metrics than OSRT (0.914 vs 0.958 FG-ARI).
Finally, note in \cref{tab:msneasy} that even if increasing resolution improves rendering quality (higher PSNR and SSIM) for all models, the perceptual similarity metric gets worse (higher LPIPS). This implies that when working at low resolution, LPIPS is not able to appropriately evaluate the model representation as perceptual inconsistencies from the 3D representation are hard to distinguish due to the poor quality. 
Therefore the usefulness of Ray-Patch querying increases. Reducing the computational cost of LFTs not only speeds-up training and inference, it also opens the possibility to work with more expensive loss functions rather than simple L1 or L2 losses, \eg using perceptual losses or adversarial discriminators, following current state of the art in image generation~\cite{esser2021taming,rombach2022high}.
\subsection{Stereo depth}
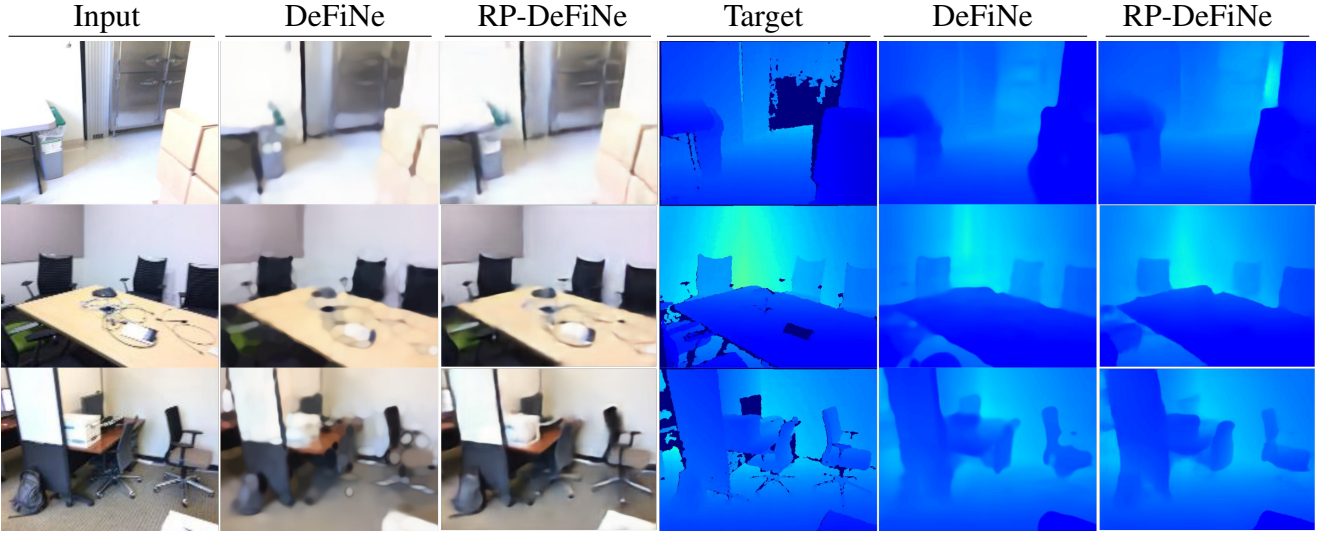
\begin{figure*}
    \centering
    \newcommand\y{200} 
    \begin{picture}(\linewidth,\y)(0,0)
    \put(0,0){\includegraphics[width=\linewidth]{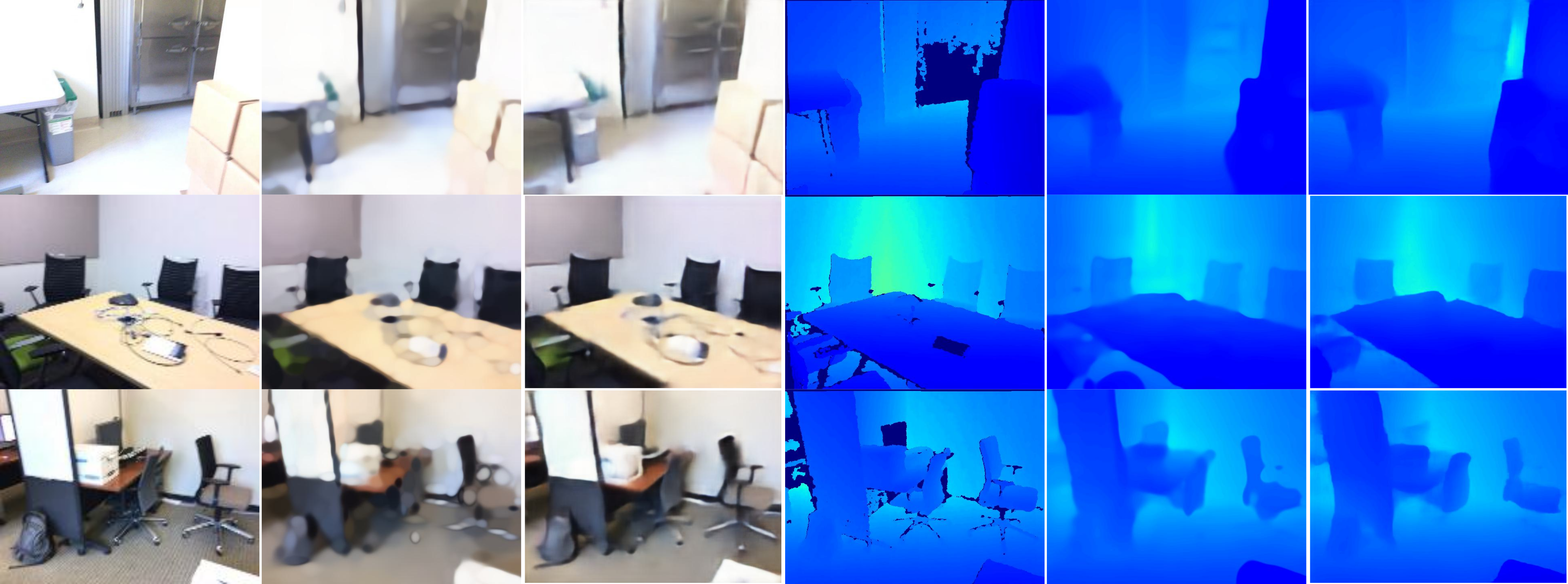}}
    \put(27, \numexpr\y-9){\fontsize{12}{0}\selectfont Input}
    \put(00, \numexpr\y-13){ \line(1, 0){78} }
    \put(107, \numexpr\y-9){\fontsize{12}{0}\selectfont DeFiNe}
    \put(82, \numexpr\y-13){ \line(1, 0){78} }
    
    \put(180, \numexpr\y-9){\fontsize{12}{0}\selectfont RP-DeFiNe}
    \put(165, \numexpr\y-13){ \line(1, 0){78} }
    
    \put(273, \numexpr\y-9){\fontsize{12}{0}\selectfont Target}
    \put(249, \numexpr\y-13){ \line(1, 0){77} }
    \put(352, \numexpr\y-9){\fontsize{12}{0}\selectfont DeFiNe}
    \put(331, \numexpr\y-13){ \line(1, 0){77} }
    \put(424, \numexpr\y-9){\fontsize{12}{0}\selectfont RP-DeFiNe}
    \put(415, \numexpr\y-13){ \line(1, 0){77} }
    
    \end{picture}
    
    \caption{\textbf{Stereo depth results on ScanNet.} Given two input images, models decode both input images and an estimation of corresponding depth maps. Note how our \(k=16\) Ray-Patch querying (RP-DeFiNe) generates sharper edges in both RGB and depth images.}
    \label{fig:scannet}
\end{figure*}
Based on the results of the previous section, the integration with DeFiNe to render at \(480\times640\) has be done with \(k=16\).  This value is chosen to have \(n_q \text{ close to }n_{kv}\), improving the computation efficiency without compressing too much the information.
Our results shows that Ray-Patch improves the convergence of the model. This configuration not only reduces the computational cost, but also improves the view reconstruction and depth stereo estimation in all metrics reported in \cref{tab:scannet}. 
Despite taking as input a \(128\times192\) image, reconstructions are closer to the \(480\times640\) target output recovering a similar quality while DeFiNe's look diffused and blurred (see \cref{fig:scannet}).
It can also be observed how the estimated depth is smoother, with less abrupt changes, while still preserving clear depth discontinuities.
Regarding the computation, the evaluated configuration reduces FLOPs \(\times10\), and increases rendering prediction of novel depth maps from 7 frames per second to 208. 
\begin{table}
\centering
\resizebox{0.8\linewidth}{!}{
\begin{tabular}{lccc}
\hline
                    & \multirow{2}{*}{DeFiNe} & RP-DeFiNe &\\ \cline{3-3}
                    & & k=16  & \\ \cline{2-3} 
 \(\uparrow\) PSNR &  23.46 & \textbf{24.54}  &\\
 \(\uparrow\) SSIM & 0.783 & \textbf{0.801}  &\\ 
\(\downarrow\) LPIPS & 0.495 & \textbf{0.453}  &\\ \hline
 \(\downarrow\)RMSE & 0.275  & \textbf{0.263}   &\\
 \(\downarrow\)Abs.Rel & 0.108  & \textbf{0.103} & \\
 \(\downarrow\)Sq.Rel  & 0.053  & \textbf{0.050} & \\\hline
\(\downarrow\) Giga FLOPs & 801 & \textbf{81} & \\
\(\uparrow\) Rendering speed & 7 fps & \textbf{208 fps} \\ \hline
\end{tabular}}
\caption{\textbf{Quantitative results on ScanNet.} Evaluation of stereo depth and RGB rendering on a realistic dataset. The integration of a Ray-Patch decoder with patch size \(k=16\) increases rendering speed by 2 orders of magnitude, while also outperforming DeFiNe's rendering and depth metrics.}
\label{tab:scannet}
\end{table}
\section{Limitations}
\label{sec:limitation}
Our proposed decoder reduces the complexity problem of decoding images with Transformers. Despite that, we cannot decode single pixels and performance may depend on choosing an appropriate patch size.
As a simple heuristic to choose the patch, we propose to keep \(n_q \sim n_{kv}\), as it has been shown that 1) rendering speed saturates for bigger patches, and 2) too much compression reduces decoding performance. Nevertheless, hyper-parameter tuning may be needed to find the best patch size for each model.
Regarding unsupervised segmentation, we observed that RP-OSRT has fallen into a tessellation failure mode, already observed by Sajjadi \cite{sajjadi2022object} \etal. This failure is dependent on architectural choices, and further experimentation would be required to address it.
Also note that we have only evaluated square patches. Nevertheless, our method could also be used with rectangular patches to obtain an intermediate number of queries.
Finally, notice that the Ray-Patch does not attempt to solve the base attention's quadratic cost scaling. Rather, its focus on reducing the number of queries makes it compatible with other less expensive alternatives to vanilla attention \cite{xiong2021nystromformer, tay2020long, zaheer2020big,dao2022flashattention, dao2023flashattention2}.
\section{Conclusion}
\label{sec:conclusion}
In this paper we propose Ray-Patch querying, which reduces significantly the cost associated to Light Field Transformer's decoder. 
Our Ray-Patch does not only reduce significantly the training time and improve convergence, but could also be generalized to different tasks using transformers to decode scenes.
We validate experimentally our approach and its benefits by integrating it into three recent LFT models for two different tasks and in two different datasets.
The models with our Ray-Patch querying match or even outperform the baseline models in photometric and depth metrics, while at the same time reducing the computation and memory load in one and two orders of magnitude respectively.
In addition, this is achieved with a minimum modification to the implementation of given baselines. 
Reducing the computational footprint of LFTs is essential for to continue its development and for deployment in constrained platforms such as mobile devices or robots, in the same line than works such as \cite{muller2022instant,wang2022yolov7} did for other architectures and tasks. 
{
    \small
    \bibliographystyle{ieeenat_fullname}
    \bibliography{main}
}
\clearpage
\setcounter{page}{1}
\maketitlesupplementary
\section{Model and training details}
All models were trained using Pytorch 2 and Pytorch Lightning on a distributed set-up comprising 4 Nvidia Tesla V100 GPUs. We intend to release both the code and checkpoints upon acceptance.
\paragraph{Convolutional Decoder with Ray-Patch Querying.} 
The convolutional decoder \(\mathcal{D}_{CNN}\) is based on GIRAFFE's decoder \cite{niemeyer2021giraffe}. It is composed of a concatenation of upsampling blocks, each incorporating preliminary outputs. The main block consist of nearest neighbour up-sampling, a convolutional layer, batch normalization, and a leaky ReLU activation function. Starting with 128 channels, each block doubles the feature map size while halving the channel count. For a patch \(k = 8 = 2^3\) there will be 3 up-sampling blocks.
To enhance the multiscale resolution, a convolutional layer preceding each block generates a preliminary output with $c$-channels. This output is added to the preliminary output from the preceding block, and is then up-scaled doubling its dimensions.
Subsequently, a final convolutional layer transforms the output from the last block into a $c$-channel result, which is added to the preliminary output to generate the ultimate output \(\hat{I}_t\).

Nevertheless, it is with noting that Ray-Patch querying could be employed with alternative up-sampling decoders, \eg a learned up-sampling decoder~\cite{shi2016real} or an attention up-sampler~\cite{esser2021taming,rombach2022high}.
\paragraph{Implementation of SRT and OSRT.} SRT and OSRT trained models were based on the implementation by \href{https://github.com/stelzner/osrt}{K. Stelzner}.
The original implementations employed batch size of 256 on 64 TPUv2~\cite{sajjadi2022object}. However, when we reduced the batch size to accommodate our hardware limitations, we observed a decline in the models' stability and convergence rate.
To address this issue, we introduced a batch normalization~\cite{ioffe2015batch} layer after each convolutional layer in the encoder. This adjustment led to improved convergence and performance, as evidenced in \cref{tab:msneasy_bn}.
Finally, both models were trained querying 9600 rays at each optimization step.
\paragraph{MSN-Easy experiments.} 
\begin{table}
\centering
\resizebox{\linewidth}{!}{
\begin{tabular}{lcccrcccrcr}
\hline
      & \multicolumn{9}{c}{MSN-Easy}                      &  \\ \cline{2-10} 
      & \multicolumn{3}{c}{with Batch Norm} & &  \multicolumn{5}{c}{w/o Batch Norm}  &  \\ \cline{2-4} \cline{6-10} 
      & \multirow{2}{*}{SRT} & \multicolumn{2}{c}{RP-SRT} && \multirow{2}{*}{SRT} & \multicolumn{2}{c}{RP-SRT} && \multirow{2}{*}{SRT} &\\ \cline{3-4} \cline{7-8}
      &                   & k = 2   & k = 4 & &   & k = 2  & k = 4  && & \\ \cline{2-4} \cline{6-10}   
      ~~~Batch        & \multicolumn{3}{c}{32} && \multicolumn{3}{c}{32} && 256&\\\cline{2-4} \cline{6-8} \cline{10-10} 
\(\uparrow\) PSNR  &  30.98 & 31.16 & 30.92 &&  29.549 & 29.576 & 29.576 &&29.32 &\\
\(\uparrow\) SSIM &  0.903 & 0.906 & 0.901 &&  0.875 & 0.875 & 0.875 &&0.876 &  \\ 
\(\downarrow\) LPIPS &  0.173 & 0.163 & 0.175 &&  0.237 & 0.230 & 0.224 &&0.200& \\\hline
\end{tabular}}
\caption{\textbf{Quantitative results of using Batch Normalization on MSN-Easy at \(\mathbf{60\times90}\) resolution with batch 32.} For the three different configurations using normalization after the convolutional layers speeds-up convergence and elevates the plateu value.}
\label{tab:msneasy_bn}
\end{table}
SRT and RP-SRT were trained using a batch size of 32, while OSRT and RP-OSRT employed a batch size of 64.
OSRT and RP-OSRT followed the original training regime~\cite{sajjadi2022object}. They were trained with an initial learning rate \(1\times 10^{-4}\), which linearly decayed to \(1.6\times 10^{-5}\) over 4M steps, also incorporating a warm-up phase of 2.5k steps. 
On the other hand, for SRT and RP-SRT we adapted the training schedule to accelerate convergence with a linear decay to \(4\times 10^{-5}\) over 300k steps.
Since all models had reached a plateau by 300k steps, we concluded the training and evaluated the best checkpoint for each model.
It is important to emphasize that the experiments are focused to compare a baseline model to its adapted version integrating Ray-Patch querying. Therefore, the different training schedules for SRT and OSRT do not exert any influence on the results and conclusions of the proposed method.
\paragraph{Scannet experiments.}
DeFiNe was implemented following the original paper \cite{guizilini2022depth}. 
Both DeFiNe and RP-DeFiNe were trained using virtual cameras projection and canonical jittering where projection noise was set to \(\sigma_v=0.25\) and canonical jittering noise was set to \(\sigma_t = \sigma_r=0.1\). Regarding the training loss, we adopted DeFiNe training loss \[\mathcal{L} = \mathcal{L}_d + \lambda_{rgb} \mathcal{L}_{rgb} + \lambda_{v}\left( \mathcal{L}_{d,v} + \lambda_{rgb} \mathcal{L}_{rgb,v}\right),\] where sub-index \(v\) is for virtual cameras, with \(\lambda_{rgb} = 5.0 \) and \(\lambda_{v} = 0.5\).
We employed AdamW optimizer\cite{AdamW} with \(\beta_1 = 0.99\), \(\beta_2=0.999\), weight decay \(w=10^-4\), and an initial learning rate of \(2\times 10^{-4}\). We trained for 200 epochs~(600k steps) halving the learning rate every 80 epochs. We did not fine tune the model at higher accuracy.
To ensure convergence stability, 1) we use gradient clipping with norm 1; and 2) we trained with a batch size of 16 and gradient accumulation 2 to simulate a batch size of 32. 
Given the \(128\times192\) stereo input images, RP-DeFiNe was trained to directly generate two \(480\times640\) output images. Instead DeFiNe, was trained querying \(32768\) of the total rays due to accommodate memory constraints. 
\section{Additional images}
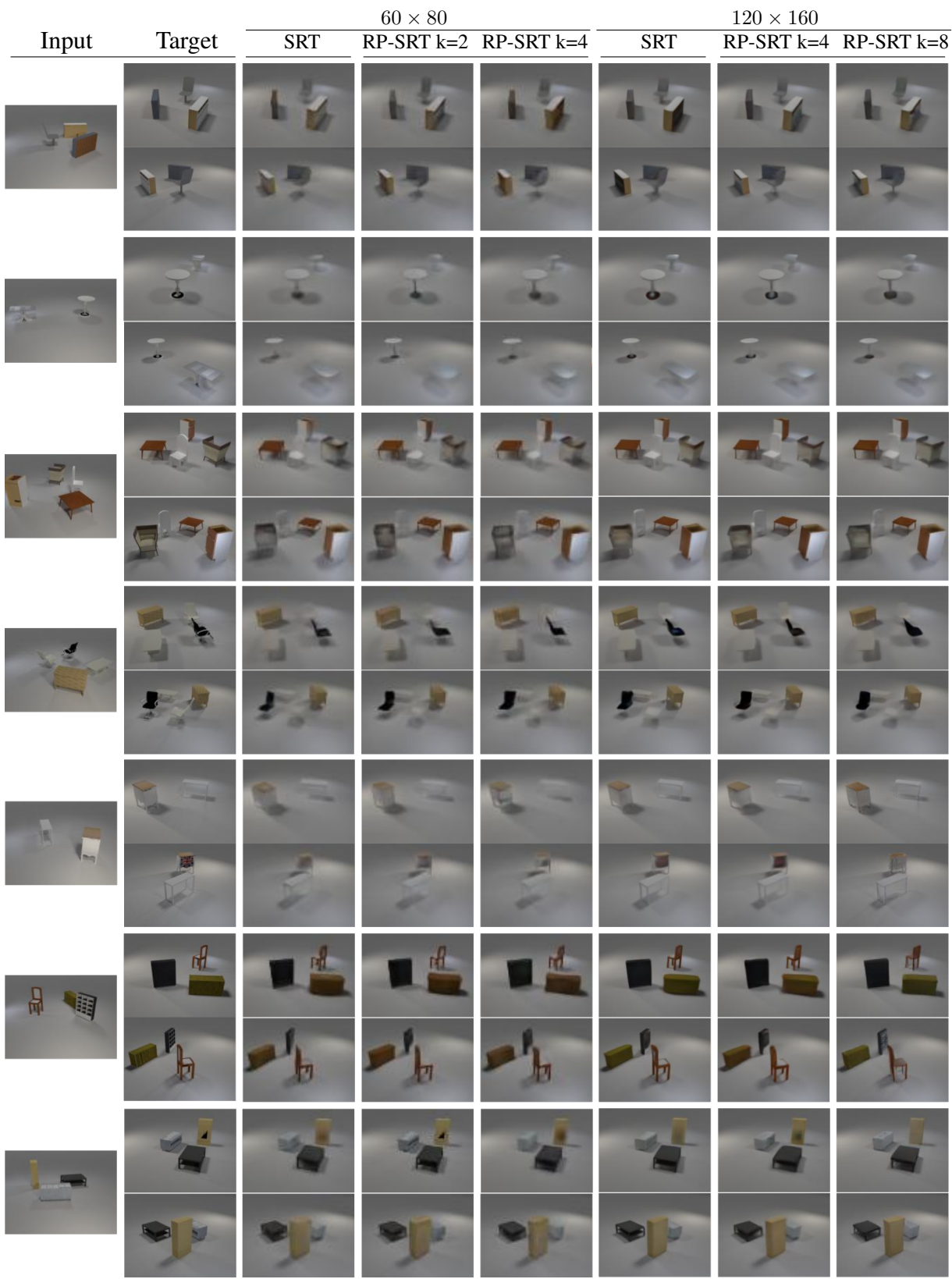
\begin{figure*}
    \centering
    \newcommand\y{606}
    \begin{picture}(\linewidth,\y)(0,0)
    \put(0,0){\includegraphics[width=0.92\linewidth]{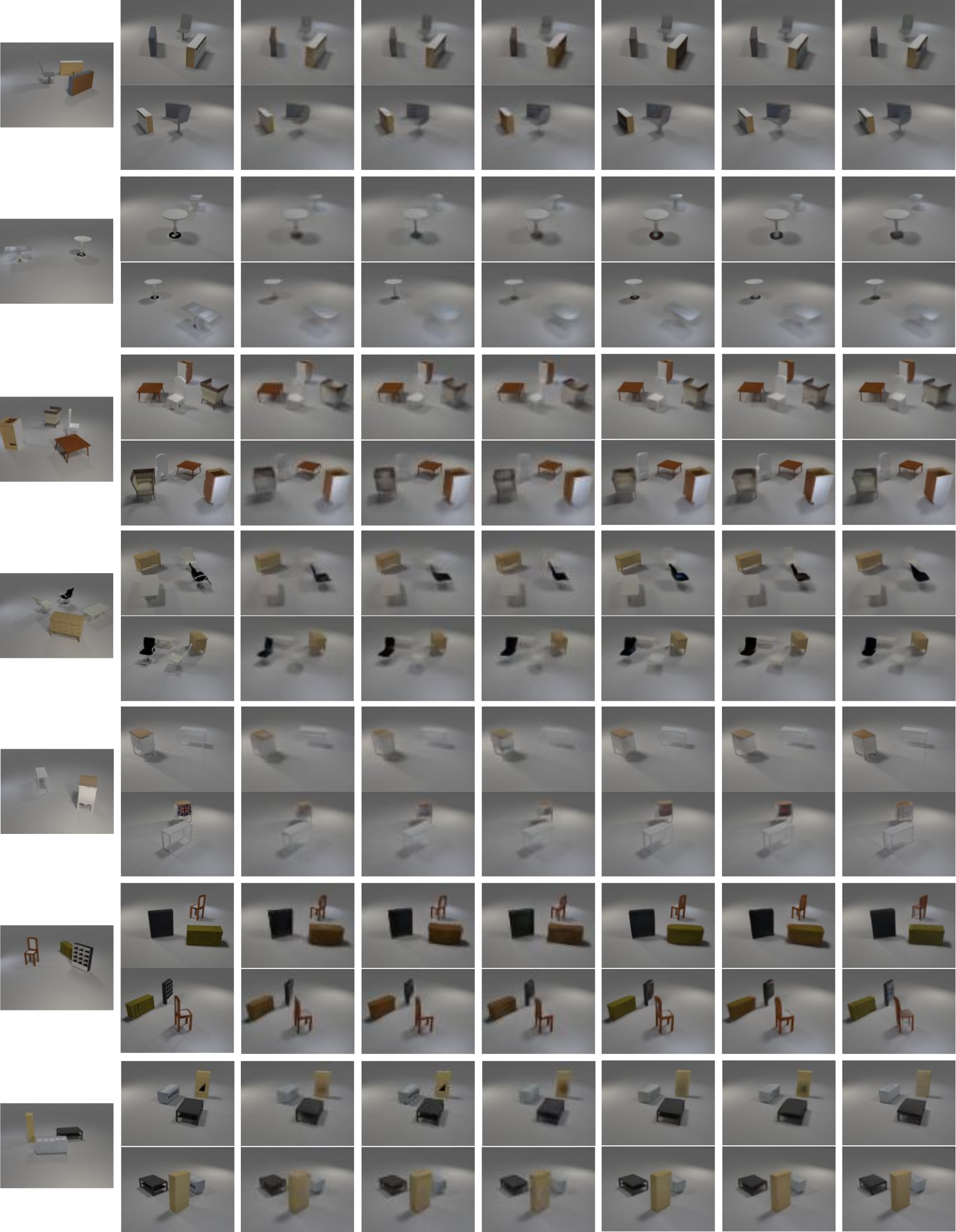}}
    \put(17, \numexpr\y-10){\fontsize{12}{0}\selectfont Input}
    \put(73, \numexpr\y-10){\fontsize{12}{0}\selectfont Target}
    \put(0, \numexpr\y-14){ \line(1, 0){109} }

    \put(182, \numexpr\y+2){\fontsize{10}{0}\selectfont \(60\times80\)}
    \put(114, \numexpr\y-1){ \line(1, 0){167} }
    \put(352, \numexpr\y+2){\fontsize{10}{0}\selectfont \(120\times160\)}
    \put(284, \numexpr\y-1){ \line(1, 0){172} }
    
    \put(135, \numexpr\y-10){\fontsize{10}{0}\selectfont SRT}
    \put(114, \numexpr\y-14){ \line(1, 0){51} }
    
    \put(173, \numexpr\y-10){\fontsize{10}{0}\selectfont RP-SRT k=2}
    \put(231, \numexpr\y-10){\fontsize{10}{0}\selectfont RP-SRT k=4}
    \put(171, \numexpr\y-14){ \line(1, 0){109} }
    
    \put(308, \numexpr\y-10){\fontsize{10}{0}\selectfont SRT}
    \put(286, \numexpr\y-14){ \line(1, 0){51} }
    
    \put(348, \numexpr\y-10){\fontsize{10}{0}\selectfont RP-SRT k=4}
    \put(406, \numexpr\y-10){\fontsize{10}{0}\selectfont RP-SRT k=8}
    \put(344, \numexpr\y-14){ \line(1, 0){109} }
    
    \end{picture}
    
    \caption{\textbf{Novel view synthesis results on MSN-Easy for SRT and RP-SRT.}}
    \label{fig:msn_easy_e1}
\end{figure*}

\begin{figure*}
    \centering
    \newcommand\y{615} 
    \begin{picture}(\linewidth,\y)(0,0)
    \put(1,0){\includegraphics[width=0.97\linewidth]{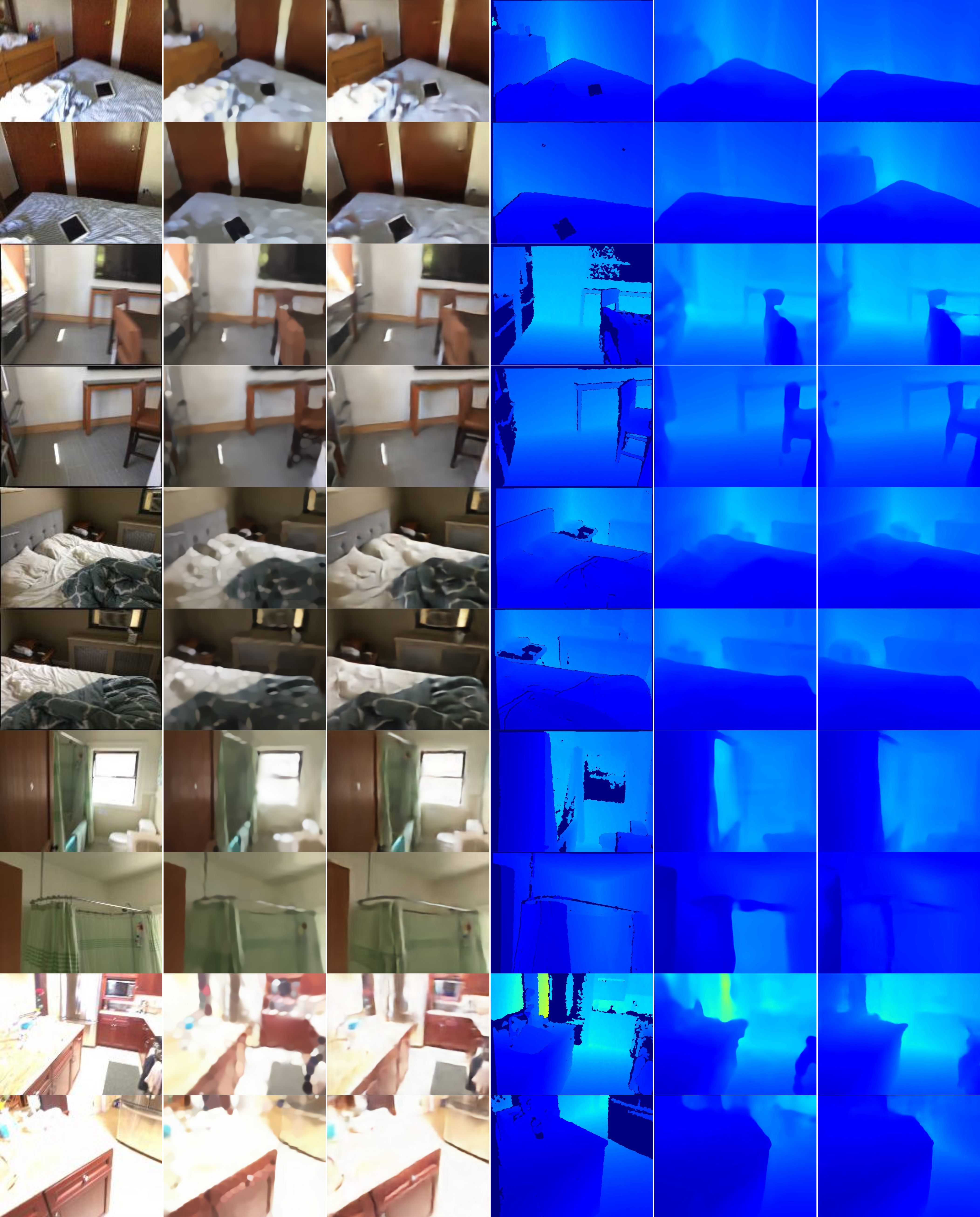}}
    \put(24, \numexpr\y-9){\fontsize{12}{0}\selectfont Input}
    \put(00, \numexpr\y-13){ \line(1, 0){75} }
    \put(104, \numexpr\y-9){\fontsize{12}{0}\selectfont DeFiNe}
    \put(79, \numexpr\y-13){ \line(1, 0){75} }
    
    \put(176, \numexpr\y-9){\fontsize{12}{0}\selectfont RP-DeFiNe}
    \put(161, \numexpr\y-13){ \line(1, 0){75} }
    
    \put(267, \numexpr\y-9){\fontsize{12}{0}\selectfont Target}
    \put(242, \numexpr\y-13){ \line(1, 0){74} }
    \put(343, \numexpr\y-9){\fontsize{12}{0}\selectfont DeFiNe}
    \put(323, \numexpr\y-13){ \line(1, 0){74} }
    \put(415, \numexpr\y-9){\fontsize{12}{0}\selectfont RP-DeFiNe}
    \put(405, \numexpr\y-13){ \line(1, 0){74} }
    
    \end{picture}
    
    \caption{\textbf{Stereo depth results on ScanNet.} }    \label{fig:scannet_e2}
\end{figure*}

Additional results for MSN-Easy and ScanNet can be observed in \cref{fig:msn_easy_e1,fig:msn_easy_e2} and \cref{fig:scannet_e1,fig:scannet_e2}.
\onecolumn
\begin{figure*}
    \centering
    \newcommand\y{616}
    \begin{picture}(\linewidth,\y)(0,0)
    \put(58,0){\includegraphics[width=0.7\linewidth]{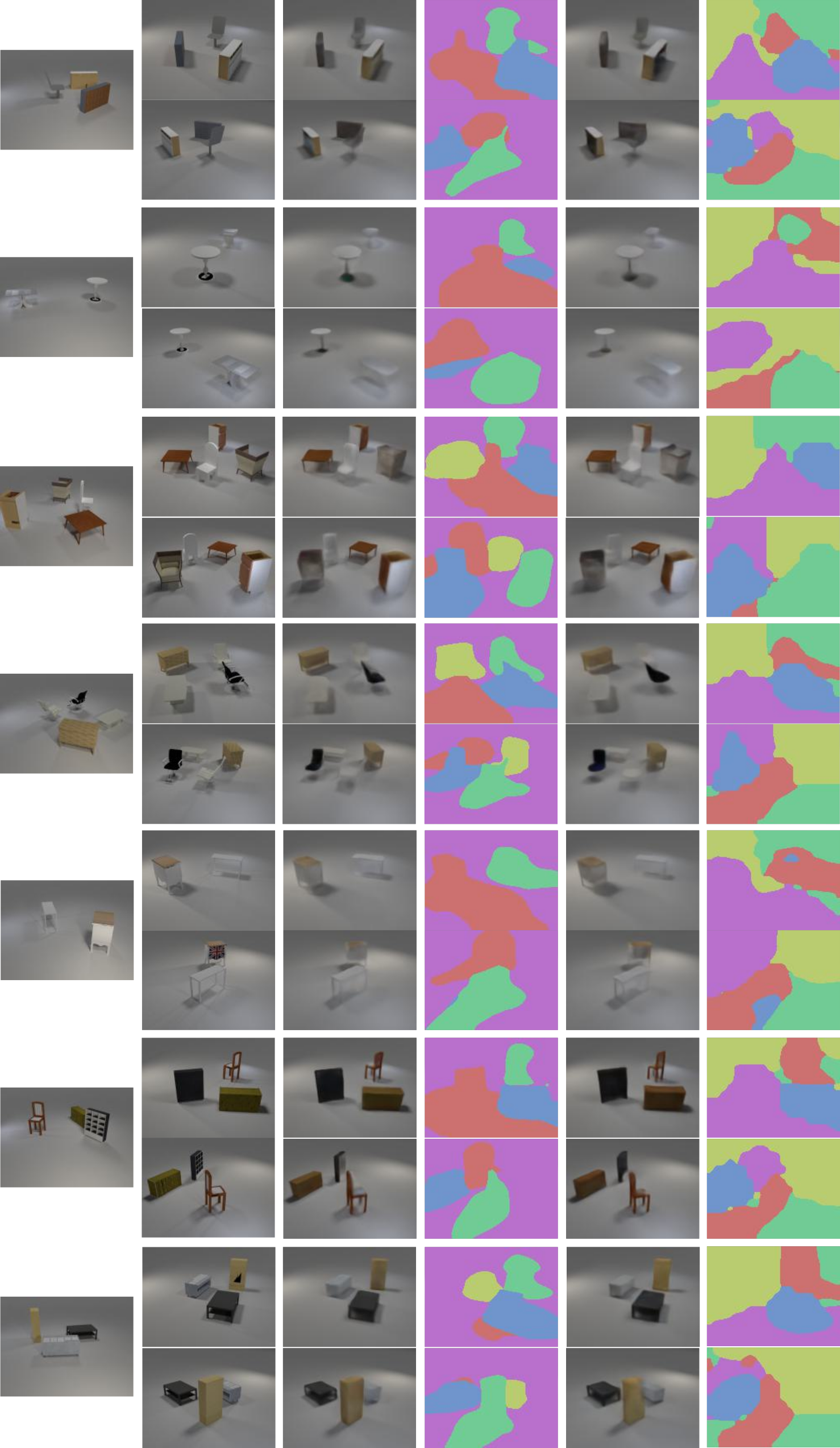}}
    \put(75, \numexpr\y-10){\fontsize{12}{0}\selectfont Input}
    \put(131, \numexpr\y-10){\fontsize{12}{0}\selectfont Target}
    \put(58, \numexpr\y-14){ \line(1, 0){112} }

    \put(216, \numexpr\y-10){\fontsize{10}{0}\selectfont SRT}
    \put(175, \numexpr\y-14){ \line(1, 0){112} }
    
    \put(325, \numexpr\y-10){\fontsize{10}{0}\selectfont RP-SRT k=8}
    \put(292, \numexpr\y-14){ \line(1, 0){112} }

    \end{picture}
    
    \caption{\textbf{Novel view synthesis results with segmentation maps on MSN-Easy for OSRT and RP-OSRT.}}
    \label{fig:msn_easy_e2}
\end{figure*}
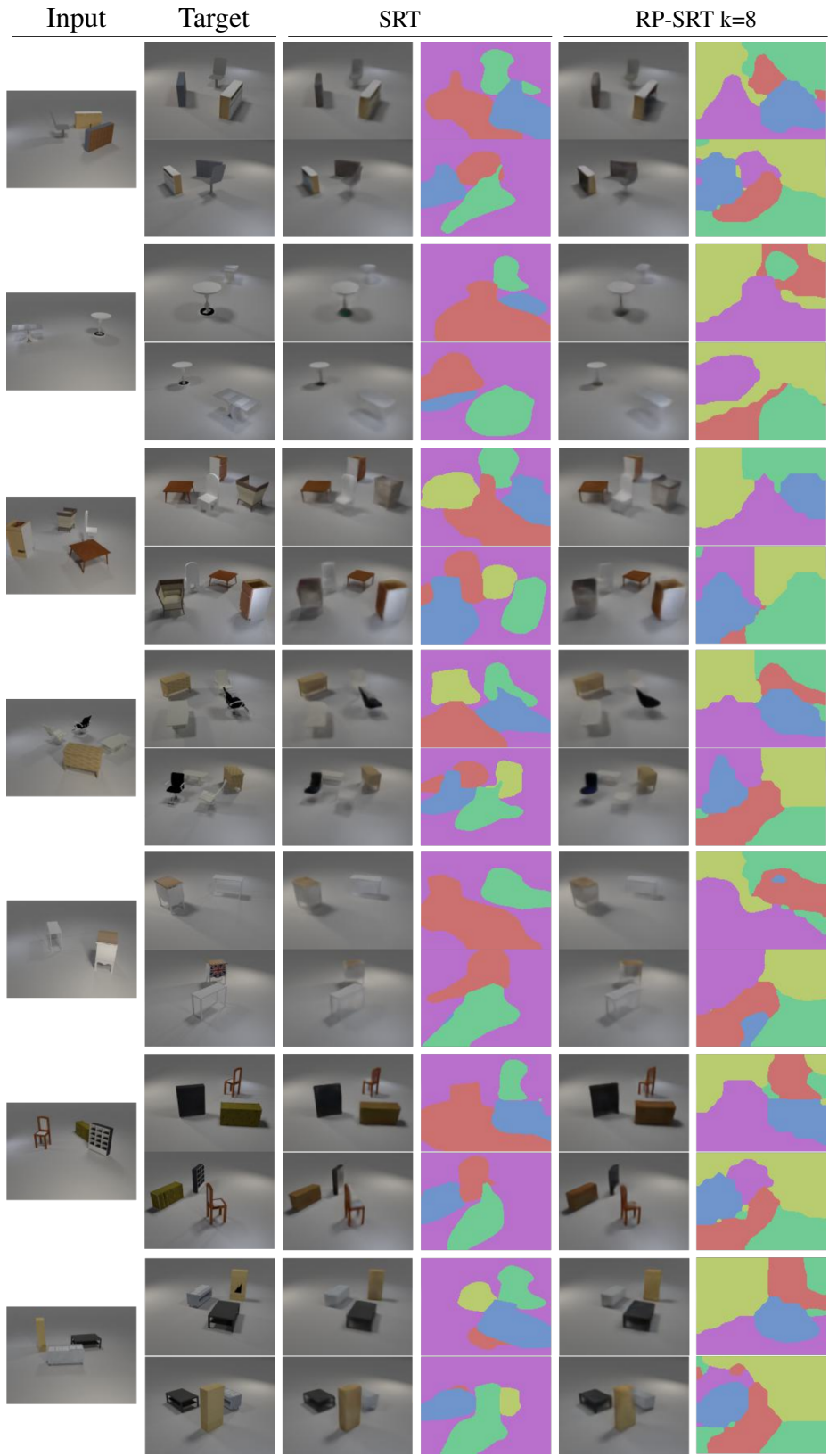
\begin{figure*}
    \centering
    \newcommand\y{615} 
    \begin{picture}(\linewidth,\y)(0,0)
    \put(1,0){\includegraphics[width=0.97\linewidth]{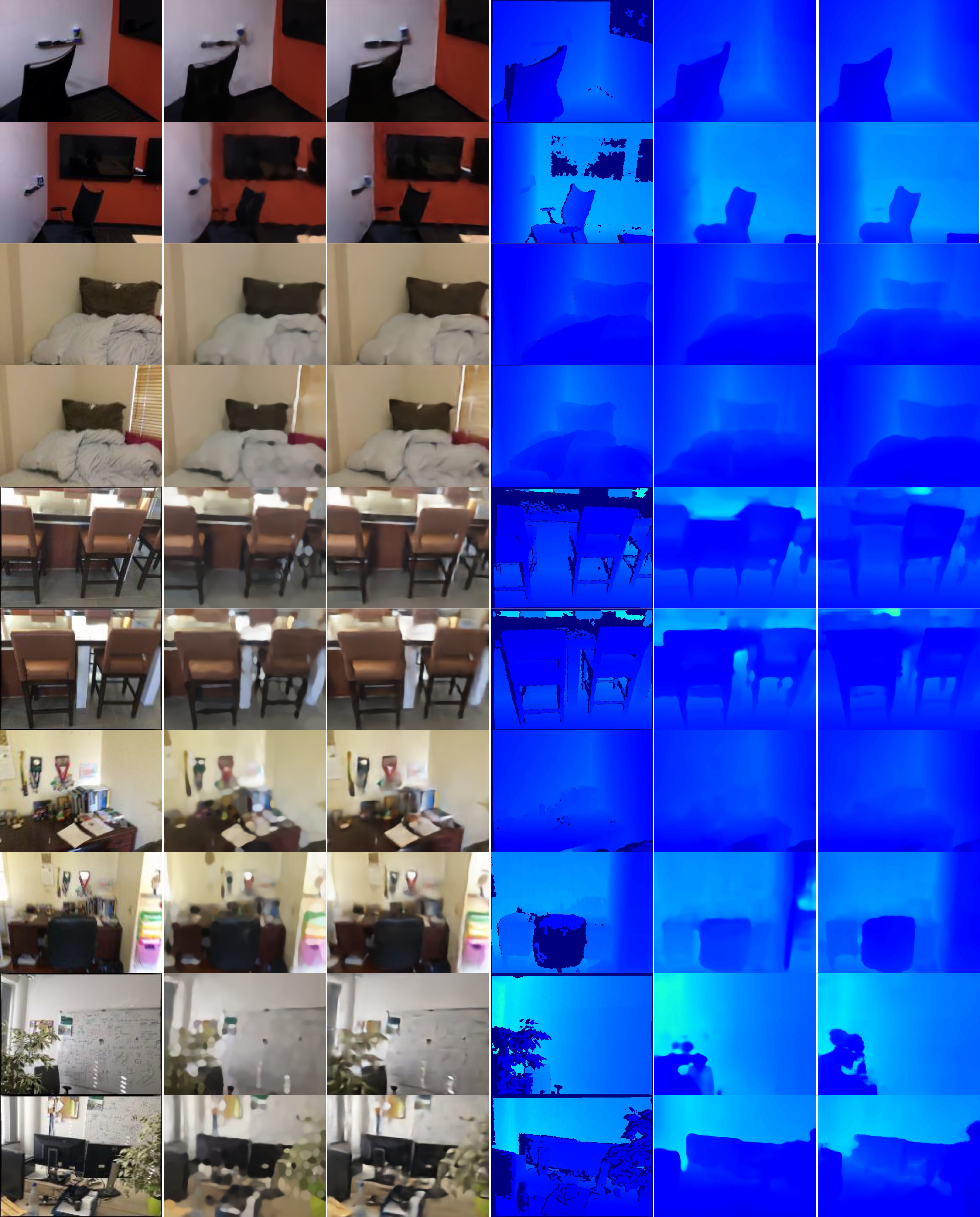}}
    \put(24, \numexpr\y-9){\fontsize{12}{0}\selectfont Input}
    \put(00, \numexpr\y-13){ \line(1, 0){75} }
    \put(104, \numexpr\y-9){\fontsize{12}{0}\selectfont DeFiNe}
    \put(79, \numexpr\y-13){ \line(1, 0){75} }
    
    \put(176, \numexpr\y-9){\fontsize{12}{0}\selectfont RP-DeFiNe}
    \put(161, \numexpr\y-13){ \line(1, 0){75} }
    
    \put(267, \numexpr\y-9){\fontsize{12}{0}\selectfont Target}
    \put(242, \numexpr\y-13){ \line(1, 0){74} }
    \put(343, \numexpr\y-9){\fontsize{12}{0}\selectfont DeFiNe}
    \put(323, \numexpr\y-13){ \line(1, 0){74} }
    \put(415, \numexpr\y-9){\fontsize{12}{0}\selectfont RP-DeFiNe}
    \put(405, \numexpr\y-13){ \line(1, 0){74} }
    
    \end{picture}
    
    \caption{\textbf{Stereo depth results on ScanNet.} }    \label{fig:scannet_e1}
\end{figure*}
\clearpage


\end{document}